\documentclass[sigconf]{acmart}
\AtBeginDocument{%
  \providecommand\BibTeX{{%
    \normalfont B\kern-0.5em{\scshape i\kern-0.25em b}\kern-0.8em\TeX}}}

\copyrightyear{2022}
\acmYear{2022}
\setcopyright{rightsretained}
\acmConference[GECCO '22 Companion]{Genetic and Evolutionary
Computation Conference Companion}{July 9--13, 2022}{Boston, MA, USA}
\acmBooktitle{Genetic and Evolutionary Computation Conference
Companion (GECCO '22 Companion), July 9--13, 2022, Boston, MA, USA}
\acmDOI{10.1145/3520304.3534073}
\acmISBN{978-1-4503-9268-6/22/07}

\setcopyright{none}
\renewcommand\footnotetextcopyrightpermission[1]{} 
\begin{document}

\title{Highlights of Semantics in Multi-objective Genetic Programming}






\author{Edgar Galv\'an}
\authornote{Main and corresponding author.}
\email{edgar.galvan@mu.ie}

\affiliation{
\institution{Dept. of CS, Hamilton Institute, IVI, Lero,  Naturally Inspired Comp. Res. Group, Maynooth University}
  \country{Ireland}
}
\author{Leonardo Trujillo}
\email{leonardo.trujillo@tectijuana.edu.mx}
\affiliation{%
  \institution{Tecnol\'ogico Nacional de M\'exico/IT de Tijuana, Tijuana, BC}
  \country{M\'exico}
}

\author{Fergal Stapleton}
\email{fergal.stapleton.2020@mumail.ie}
\affiliation{
\institution{Dept. of CS, Hamilton Institute,  Naturally Inspired Comp. Res. Group, Maynooth University}
  \country{Ireland}
}







\begin{abstract}




Semantics is a growing area of research in Genetic programming (GP) and refers to the behavioural output of a Genetic Programming individual when executed. Thus far, the majority of works relating to semantics have been focused within a single objective context and to date there has been little research into the use of semantics in Multi-objective GP (MOGP). This research expands upon the current understanding of semantics by proposing a new approach: Semantic-based
Distance as an additional criteriOn (SDO) to use in a MOGP context. Our work included an expansive analysis of the GP in terms of performance and diversity metrics, using two additional semantic based approaches, namely Semantic Similarity-based Crossover (SCC)
and Semantic-based Crowding Distance (SCD). Each approach is integrated into two evolutionary multi-objective (EMO) frameworks: Non-dominated Sorting Genetic
Algorithm II (NSGA-II) and the Strength Pareto Evolutionary Algorithm 2 (SPEA2), and along with the three semantic approaches, the canonical form of NSGA-II and SPEA2 are rigorously compared. To discuss some limitations of the SDO approach we also do a comparison with a decompositional based framework: Multi-objective Evolutionary Algorithm with Decomposition \sloppy{(MOEA/D)}. Using highly-unbalanced binary classification datasets, we demonstrated that the newly proposed approach of SDO consistently generated more non-dominated solutions, with better diversity and improved hypervolume results.

\sloppy\textit{This Hot-off-the-Press paper summarises "Semantics in Multi-objective Genetic Programming" by Edgar Galván, Leonardo Trujillo and Fergal Stapleton, published in the journal of Applied Soft Computing 2022~\cite{GALVAN2022108143}, \url{https://doi.org/10.1016/j.asoc.2021.108143}.}

\end{abstract}

\maketitle
\pagestyle{plain}

\section{Approaches, Contributions \& Results}

Genetic Programming (GP) is a long established paradigm of Evolutionary Algorithms that first came into prominence in the early 1990's~\cite{Koza:1992:GPP:138936}. Since then, researchers have sought to improve the search capabilities and robustness of GP, for example the use of dynamic fitness cases can make search more amenable in GP~\cite{DBLP:conf/gecco/LopezVST17, DBLP:conf/ae/LopezVST17}. Another emergent approach is the area of semantics. Broadly speaking semantics can be understood as the behaviour of a GP program once it has been executed on a set of fitness cases. To date, the majority of works in semantics have focused on single objective GP (SOGP). 
This work greatly expands upon previous research into semantics in MOGP~\cite{DBLP:conf/gecco/GalvanS19,Galvan-Lopez2016,Galvan_MICAI_2016,9308386, stapleton2021cec}. Specifically, we investigate three semantic based approaches using two long established and popular evolutionary multi-objective (EMO) based algorithms: the Non-dominated Sorting Genetic Algorithm
II (NSGA-II)~\cite{996017} and the Strength Pareto Evolutionary Algorithm
(SPEA2)~\cite{934438}. Additionally, we do a comparison with a decompositional EMO approach, namely the Multi-objective Evolutionary Algorithm with Decomposition (MOEA/D), to highlight some limitations of our proposed semantic-based approach.

\textit{Semantic Similarity-based Crossover} (SSC) is motivated by the seminal work of Uy et. al.~\cite{Uy2011} and was notable as an early example of semantics being applied to continuous search spaces in a SOGP context. The semantic distance metric for SSC is calculated as follows: for every input $in \in I$, where $I$ is the partial set of inputs, the absolute difference of values between parents are calculated. The semantic distance is then computed as the average of the differences. Using an Upper-Bound Semantic Similarity (UBSS) and Lower-Bound Semantic Similarity (LBSS) metric, if the distance value falls within this range, then the crossover operation is promoted. 

\begin{figure}[tb!]
   \includegraphics[width=0.75\columnwidth]{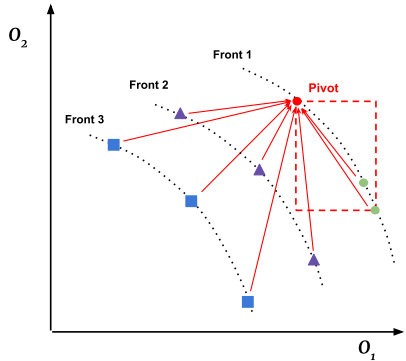}
\caption{The pivot is is selected using crowding distance (represented by red dotted rectangle). The semantic distance is calculated between all solutions in $R_t$ and the pivot. }
\label{fig:SDO_aproach}
\end{figure}

\textit{Semantic-based Crowding Distance} (SCD) is in part inspired by the crowding distance metric found in EMOs such as NSGA-II and SPEA2, and is incorporated in its place. As is the case with cannonical NSGA-II and SPEA2, the solutions are first sorted based upon the strengths of each solution. These strengths are determined using either dominance rank (NSGA-II) or both dominance rank and dominance count (SPEA2). The solutions are then stored in population $R_t$. After we find the non-dominated solutions of the first front (best front), and from this front we can then select an individual (pivot) from the sparsest region using the crowding distance (represented by the red dotted rectangle, see Fig.~\ref{fig:SDO_aproach}). The semantic distance is then calculated between the pivot and every individual stored in $R_t$. Eqs.~\ref{semantic:distance:one:value} and ~\ref{semantic:distance:two:values} demonstrate how the semantic distance is calculated:

\begin{equation}
\label{semantic:distance:one:value}
  d(p_j,v) = \sum_{i=1}^l 1 \text{ if }   |p(in_i) - v(in_i)| > \text{UBSS}  
\end{equation}

\begin{equation}
  \label{semantic:distance:two:values}
    d(p_j,v) = \sum_{i=1}^l 1 \text{ if } \text{LBSS} \leq |p(in_i) - v(in_i)| \leq \text{UBSS} 
  \end{equation}

\noindent where $p_j$ is an individual in $R_t$, $v$ is the pivot and $l$ is the number of fitness cases. Again, a LBSS and UBSS are used in this approach, albeit in a different manner to SSC.

\textit{Semantic-based
Distance as an additional criteriOn} (SDO) expands upon the SCD approach by using the semantic distance as an additional criterion to optimize along with the objectives $O_1$ and $O_2$, which represent the conflicting TPR and TNR of the unbalanced binary classification problem~\cite{GALVAN2022108143}. The motivation for using the semantic distance in this manner, is that the individual from the sparsest region of the non-dominated front, represents the most diverse individual from that front. Therefore, we seek to further promote individuals that are behaviourally similar to that individual. In essence, we wish to attract individuals to the regions of sparsity, offering a better spread for our approximated front.

\setlength{\belowcaptionskip}{-10pt}

\subsection*{Analysis and Contributions} 

\begin{itemize}
    
\item This work analyzed the phenotypic diversity of solutions, based upon the number of unique solutions. The SDO approach had on average, unique solution ratios ranging from $\approx$ 2.4 to 5.6 compared to NSGA-II and $\approx$ 2.7 to 6.7 compared to SPEA2 depending on the data set under consideration. Neither the SSC or SCD approach produced an advantage in terms of unqiue solutions when compared to NSGA-II. 
\item A rigorous statistical analysis compared hypervolumes for the three semantic approaches for NSGA-II and SPEA-II, each with 16 separate configurations of LBSS and UBSS values, showing that overall the SDO approach performed significantly better than the other approaches. Interestingly, though SSC was shown to have beneficial performance in SOGP this was not observed in MOGP.
\item The number of nodes demonstrated that the SDO approach actually produced more nodes than the canonical approach and the other two semantic approaches. This suggests the additional growth actually leads to an improvement in performance, which is not normally seen with bloat.

\item This work described in detail how the SDO approach improves evolutionary search. Moreover, we provide a limitation discussion that describes how and why SDO fails when incorporated into a decomposition-based approach rather than the Pareto-based approaches predominantly used in our analysis.

\end{itemize}






\bibliographystyle{ACM-Reference-Format}
\bibliography{ref}


\begin{thebibliography}{12}


\ifx \showCODEN    \undefined \def \showCODEN     #1{\unskip}     \fi
\ifx \showDOI      \undefined \def \showDOI       #1{#1}\fi
\ifx \showISBNx    \undefined \def \showISBNx     #1{\unskip}     \fi
\ifx \showISBNxiii \undefined \def \showISBNxiii  #1{\unskip}     \fi
\ifx \showISSN     \undefined \def \showISSN      #1{\unskip}     \fi
\ifx \showLCCN     \undefined \def \showLCCN      #1{\unskip}     \fi
\ifx \shownote     \undefined \def \shownote      #1{#1}          \fi
\ifx \showarticletitle \undefined \def \showarticletitle #1{#1}   \fi
\ifx \showURL      \undefined \def \showURL       {\relax}        \fi
\providecommand\bibfield[2]{#2}
\providecommand\bibinfo[2]{#2}
\providecommand\natexlab[1]{#1}
\providecommand\showeprint[2][]{arXiv:#2}

\bibitem[\protect\citeauthoryear{Bleuler, Brack, Thiele, and Zitzler}{Bleuler
  et~al\mbox{.}}{2001}]%
        {934438}
\bibfield{author}{\bibinfo{person}{Stefan Bleuler}, \bibinfo{person}{Martin
  Brack}, \bibinfo{person}{Lothar Thiele}, {and} \bibinfo{person}{Eckart
  Zitzler}.} \bibinfo{year}{2001}\natexlab{}.
\newblock \showarticletitle{Multiobjective genetic programming: reducing bloat
  using SPEA2}. In \bibinfo{booktitle}{\emph{Proceedings of the 2001 Congress
  on Evolutionary Computation (IEEE Cat. No.01TH8546)}},
  Vol.~\bibinfo{volume}{1}. \bibinfo{pages}{536--543 vol. 1}.
\newblock
\urldef\tempurl%
\url{https://doi.org/10.1109/CEC.2001.934438}
\showDOI{\tempurl}


\bibitem[\protect\citeauthoryear{Deb, Pratap, Agarwal, and Meyarivan}{Deb
  et~al\mbox{.}}{2002}]%
        {996017}
\bibfield{author}{\bibinfo{person}{Kalyanmoy Deb}, \bibinfo{person}{Amrit
  Pratap}, \bibinfo{person}{Sameer Agarwal}, {and} \bibinfo{person}{TAMT
  Meyarivan}.} \bibinfo{year}{2002}\natexlab{}.
\newblock \showarticletitle{A fast and elitist multiobjective genetic
  algorithm: NSGA-II}.
\newblock \bibinfo{journal}{\emph{IEEE Transactions on Evolutionary
  Computation}} \bibinfo{volume}{6}, \bibinfo{number}{2}
  (\bibinfo{year}{2002}), \bibinfo{pages}{182--197}.
\newblock
\urldef\tempurl%
\url{https://doi.org/10.1109/4235.996017}
\showDOI{\tempurl}


\bibitem[\protect\citeauthoryear{Galv{\'{a}}n and Schoenauer}{Galv{\'{a}}n and
  Schoenauer}{2019}]%
        {DBLP:conf/gecco/GalvanS19}
\bibfield{author}{\bibinfo{person}{Edgar Galv{\'{a}}n} {and}
  \bibinfo{person}{Marc Schoenauer}.} \bibinfo{year}{2019}\natexlab{}.
\newblock \showarticletitle{Promoting semantic diversity in multi-objective
  genetic programming}. In \bibinfo{booktitle}{\emph{Proceedings of the Genetic
  and Evolutionary Computation Conference, {GECCO} 2019, Prague, Czech
  Republic, July 13-17, 2019}}, \bibfield{editor}{\bibinfo{person}{Anne Auger}
  {and} \bibinfo{person}{Thomas St{\"{u}}tzle}} (Eds.).
  \bibinfo{publisher}{{ACM}}, \bibinfo{pages}{1021--1029}.
\newblock
\showISBNx{978-1-4503-6111-8}
\urldef\tempurl%
\url{https://doi.org/10.1145/3321707.3321854}
\showDOI{\tempurl}


\bibitem[\protect\citeauthoryear{Galv{\'a}n and Stapleton}{Galv{\'a}n and
  Stapleton}{2020}]%
        {9308386}
\bibfield{author}{\bibinfo{person}{Edgar Galv{\'a}n} {and}
  \bibinfo{person}{Fergal Stapleton}.} \bibinfo{year}{2020}\natexlab{}.
\newblock \showarticletitle{Semantic-based Distance Approaches in
  Multi-objective Genetic Programming}. In \bibinfo{booktitle}{\emph{2020 IEEE
  Symposium Series on Computational Intelligence (SSCI)}}.
  \bibinfo{pages}{149--156}.
\newblock
\urldef\tempurl%
\url{https://doi.org/10.1109/SSCI47803.2020.9308386}
\showDOI{\tempurl}


\bibitem[\protect\citeauthoryear{Galv{\'a}n-L{\'o}pez, Mezura-Montes,
  Ait~ElHara, and Schoenauer}{Galv{\'a}n-L{\'o}pez et~al\mbox{.}}{2016}]%
        {Galvan-Lopez2016}
\bibfield{author}{\bibinfo{person}{Edgar Galv{\'a}n-L{\'o}pez},
  \bibinfo{person}{Efr{\'e}n Mezura-Montes}, \bibinfo{person}{Ouassim
  Ait~ElHara}, {and} \bibinfo{person}{Marc Schoenauer}.}
  \bibinfo{year}{2016}\natexlab{}.
\newblock \showarticletitle{On the Use of Semantics in Multi-objective Genetic
  Programming}. In \bibinfo{booktitle}{\emph{Parallel Problem Solving from
  Nature -- PPSN XIV: 14th International Conference, Edinburgh, UK, September
  17-21, 2016, Proceedings}}, \bibfield{editor}{\bibinfo{person}{Julia Handl}
  {et~al\mbox{.}}} (Eds.). \bibinfo{publisher}{Springer},
  \bibinfo{pages}{353--363}.
\newblock
\showISBNx{978-3-319-45823-6}
\urldef\tempurl%
\url{https://doi.org/10.1007/978-3-319-45823-6_33}
\showDOI{\tempurl}


\bibitem[\protect\citeauthoryear{Galv{\'{a}}n-L{\'{o}}pez,
  V{\'{a}}zquez{-}Mendoza, Schoenauer, and Trujillo}{Galv{\'{a}}n-L{\'{o}}pez
  et~al\mbox{.}}{2017}]%
        {DBLP:conf/gecco/LopezVST17}
\bibfield{author}{\bibinfo{person}{Edgar Galv{\'{a}}n-L{\'{o}}pez},
  \bibinfo{person}{Lucia V{\'{a}}zquez{-}Mendoza}, \bibinfo{person}{Marc
  Schoenauer}, {and} \bibinfo{person}{Leonardo Trujillo}.}
  \bibinfo{year}{2017}\natexlab{}.
\newblock \showarticletitle{Dynamic {GP} fitness cases in static and dynamic
  optimisation problems}. In \bibinfo{booktitle}{\emph{Genetic and Evolutionary
  Computation Conference, Berlin, Germany, July 15-19, 2017, Companion Material
  Proceedings}}, \bibfield{editor}{\bibinfo{person}{Peter A.~N. Bosman}} (Ed.).
  \bibinfo{publisher}{{ACM}}, \bibinfo{pages}{227--228}.
\newblock
\showISBNx{978-1-4503-4939-0}
\urldef\tempurl%
\url{https://doi.org/10.1145/3067695.3076055}
\showDOI{\tempurl}


\bibitem[\protect\citeauthoryear{Galv{\'a}n-L{\'o}pez, V{\'a}zquez-Mendoza,
  Schoenauer, and Trujillo}{Galv{\'a}n-L{\'o}pez et~al\mbox{.}}{2017}]%
        {DBLP:conf/ae/LopezVST17}
\bibfield{author}{\bibinfo{person}{Edgar Galv{\'a}n-L{\'o}pez},
  \bibinfo{person}{Lucia V{\'a}zquez-Mendoza}, \bibinfo{person}{Marc
  Schoenauer}, {and} \bibinfo{person}{Leonardo Trujillo}.}
  \bibinfo{year}{2017}\natexlab{}.
\newblock \showarticletitle{On the use of dynamic GP fitness cases in static
  and dynamic optimisation problems}. In
  \bibinfo{booktitle}{\emph{International Conference on Artificial Evolution
  (Evolution Artificielle)}}. Springer, \bibinfo{pages}{72--87}.
\newblock


\bibitem[\protect\citeauthoryear{Galv\'an-L\'opez, V\'azquez-Mendoza, and
  Trujillo}{Galv\'an-L\'opez et~al\mbox{.}}{2016}]%
        {Galvan_MICAI_2016}
\bibfield{author}{\bibinfo{person}{Edgar Galv\'an-L\'opez},
  \bibinfo{person}{Lucia V\'azquez-Mendoza}, {and} \bibinfo{person}{Leonardo
  Trujillo}.} \bibinfo{year}{2016}\natexlab{}.
\newblock \showarticletitle{Stochastic Semantic-Based Multi-Objective Genetic
  Programming Optimisation for Classification of Imbalanced Data}.
\newblock In \bibinfo{booktitle}{\emph{Advances in Soft Computing}},
  \bibfield{editor}{\bibinfo{person}{Obdulia Pichardo-Lagunas} {and}
  \bibinfo{person}{Sabino Miranda-Jim\'enez}} (Eds.).
  \bibinfo{publisher}{Springer}, Chapter~22, \bibinfo{pages}{261--272}.
\newblock


\bibitem[\protect\citeauthoryear{Galván, Trujillo, and Stapleton}{Galván
  et~al\mbox{.}}{2022}]%
        {GALVAN2022108143}
\bibfield{author}{\bibinfo{person}{Edgar Galván}, \bibinfo{person}{Leonardo
  Trujillo}, {and} \bibinfo{person}{Fergal Stapleton}.}
  \bibinfo{year}{2022}\natexlab{}.
\newblock \showarticletitle{Semantics in Multi-objective Genetic Programming}.
\newblock \bibinfo{journal}{\emph{Applied Soft Computing}}
  \bibinfo{volume}{115} (\bibinfo{year}{2022}), \bibinfo{pages}{108143}.
\newblock
\showISSN{1568-4946}
\urldef\tempurl%
\url{https://doi.org/10.1016/j.asoc.2021.108143}
\showDOI{\tempurl}


\bibitem[\protect\citeauthoryear{Koza}{Koza}{1992}]%
        {Koza:1992:GPP:138936}
\bibfield{author}{\bibinfo{person}{John~R. Koza}.}
  \bibinfo{year}{1992}\natexlab{}.
\newblock \bibinfo{booktitle}{\emph{Genetic Programming: On the Programming of
  Computers by Means of Natural Selection}}.
\newblock \bibinfo{publisher}{MIT Press}, \bibinfo{address}{Cambridge, MA,
  USA}.
\newblock
\showISBNx{0-262-11170-5}


\bibitem[\protect\citeauthoryear{Stapleton and Galván}{Stapleton and
  Galván}{2021}]%
        {stapleton2021cec}
\bibfield{author}{\bibinfo{person}{Fergal Stapleton} {and}
  \bibinfo{person}{Edgar Galván}.} \bibinfo{year}{2021}\natexlab{}.
\newblock \showarticletitle{Semantic Neighborhood Ordering in Multi-objective
  Genetic Programming based on Decomposition}. In
  \bibinfo{booktitle}{\emph{2021 IEEE Congress on Evolutionary Computation
  (CEC)}}. \bibinfo{pages}{580--587}.
\newblock
\urldef\tempurl%
\url{https://doi.org/10.1109/CEC45853.2021.9504860}
\showDOI{\tempurl}


\bibitem[\protect\citeauthoryear{Uy, Hoai, O'Neill, McKay, and
  Galv{\'a}n-L{\'o}pez}{Uy et~al\mbox{.}}{2011}]%
        {Uy2011}
\bibfield{author}{\bibinfo{person}{Nguyen~Quang Uy},
  \bibinfo{person}{Nguyen~Xuan Hoai}, \bibinfo{person}{Michael O'Neill},
  \bibinfo{person}{R.~I. McKay}, {and} \bibinfo{person}{Edgar
  Galv{\'a}n-L{\'o}pez}.} \bibinfo{year}{2011}\natexlab{}.
\newblock \showarticletitle{Semantically-based crossover in genetic
  programming: application to real-valued symbolic regression}.
\newblock \bibinfo{journal}{\emph{Genetic Programming and Evolvable Machines}}
  \bibinfo{volume}{12}, \bibinfo{number}{2} (\bibinfo{year}{2011}),
  \bibinfo{pages}{91--119}.
\newblock
\showISSN{1573-7632}
\urldef\tempurl%
\url{https://doi.org/10.1007/s10710-010-9121-2}
\showDOI{\tempurl}


\end{thebibliography}

\begin{acks}
        This publication has emanated from research conducted with the financial support of Science Foundation Ireland under Grant number 18/CRT/6049. 
\end{acks}



\end{document}